\documentclass{article}

\usepackage{microtype}
\usepackage{graphicx}
\usepackage{subfigure}
\usepackage{booktabs} 

\usepackage{hyperref}

\usepackage[accepted]{icml2024}

\usepackage{amsmath}
\usepackage{amssymb}
\usepackage{mathtools}
\usepackage{amsthm}

\usepackage[capitalize,noabbrev]{cleveref}

\theoremstyle{plain}

\theoremstyle{definition}

\theoremstyle{remark}

\usepackage[textsize=tiny]{todonotes}
\usepackage{multirow}
\usepackage{hyperref}

\icmltitlerunning{Quantum Implicit Neural Representations}

\begin{document}

\twocolumn[
\icmltitle{Quantum Implicit Neural Representations}



\icmlsetsymbol{equal}{*}

\begin{icmlauthorlist}
\icmlauthor{Jiaming Zhao}{sch1}
\icmlauthor{Wenbo Qiao}{sch1}
\icmlauthor{$\text{Peng Zhang}^{*}$}{sch2}
\icmlauthor{Hui Gao}{sch2}
\end{icmlauthorlist}
\icmlaffiliation{sch1}{School of New Media and Communication, Tianjin University, China}
\icmlaffiliation{sch2}{College of Intelligence and Computing, Tianjin University, China}

\icmlcorrespondingauthor{Peng Zhang}{pzhang@tju.edu.cn}

\icmlkeywords{Machine Learning, ICML}

\vskip 0.3in

]



\printAffiliationsAndNotice{} 

\begin{abstract}
Implicit neural representations have emerged as a powerful paradigm to represent signals such as images and sounds. This approach aims to utilize neural networks to parameterize the implicit function of the signal. However, when representing implicit functions, traditional neural networks such as ReLU-based multilayer perceptrons face challenges in accurately modeling high-frequency components of signals. Recent research has begun to explore the use of Fourier Neural Networks (FNNs) to overcome this limitation. In this paper, we propose \textbf{Q}uantum \textbf{I}mplicit \textbf{Re}presentation \textbf{N}etwork (QIREN), a novel quantum generalization of FNNs. Furthermore, through theoretical analysis, we demonstrate that QIREN possesses a quantum advantage over classical FNNs. Lastly, we conducted experiments in signal representation, image superresolution, and image generation tasks to show the superior performance of QIREN compared to state-of-the-art (SOTA) models. Our work not only incorporates quantum advantages into implicit neural representations but also uncovers a promising application direction for Quantum Neural Networks. Our code is available at \href{https://github.com/GGorMM1/QIREN}{https://github.com/GGorMM1/QIREN}.
\end{abstract}

\section{Introduction}
\begin{figure}[h]
\centering
\includegraphics[scale=0.27]{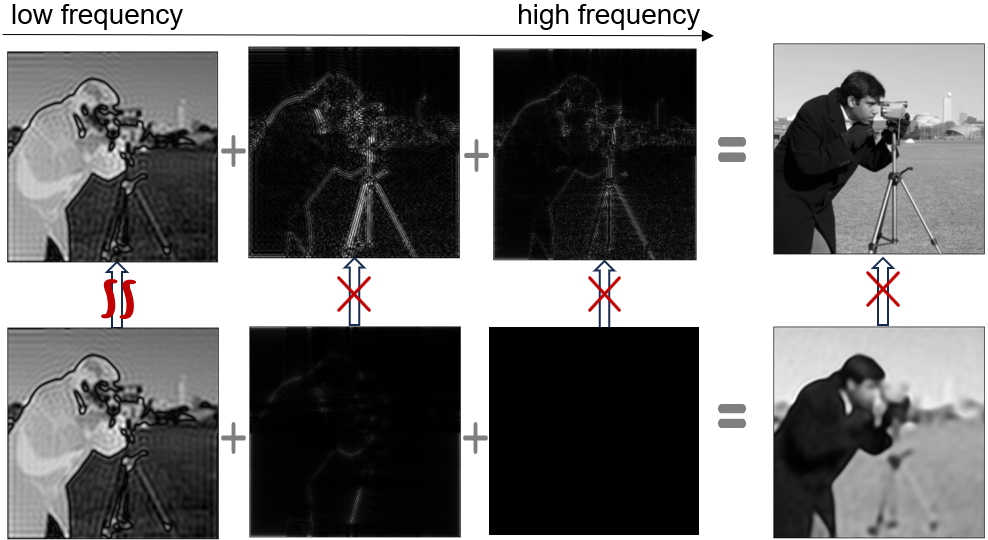}
\caption{The different frequency components of a real image (top) and the image fitted by a ReLU-based MLP (bottom).}
\label{fig1}
\end{figure}
Implicit neural representations (INRs) are a novel research field where traditional discrete grid-sampled signal representations (e.g., images as discrete grids of pixels) are replaced by continuous functions. These continuous functions are typically implicitly defined, and INRs approximate them with deep neural networks which take low-dimensional coordinates as inputs and output numerical values, such as color, amplitude, and density, at corresponding positions. INRs are not coupled to spatial resolution, thus offering improved memory efficiency and overcoming resolution limitations \cite{dupont2021coin}. Most of early research on INRs is built on ReLU-based multilayer perceptrons (MLPs) \cite{park2019deepsdf,sitzmann2019scene,jiang2020local,genova2019learning}. However, there exists a frequency principle \cite{xu2019frequency,rahaman2019spectral} that the spectrum of ReLU-based MLPs rapidly decays with increasing frequency, which leads to the limited capability of ReLU-based MLPs in modeling high-frequency components of signals, as illustrated in Figure \ref{fig1}.

On one hand, recent studies \cite{mildenhall2021nerf, sitzmann2020implicit, tancik2020fourier} have discovered that employing Fourier Neural Networks can partially address this issue. Specifically, by mapping the input using trigonometric functions or replacing the ReLU activation function with trigonometric functions, the capacity of the network to represent high-frequency components of the signal can be significantly enhanced. However, faced with increasingly complex fitting tasks in real-world applications, the classical Fourier Neural Network also requires a growing number of training parameters, which increases the demand for computing resources. It is crucial for the field of INRs and even machine learning to explore new solutions that can reduce these parameters and alleviate computational consumption.

\begin{figure*}[t]
\centering
\includegraphics[scale=0.36]{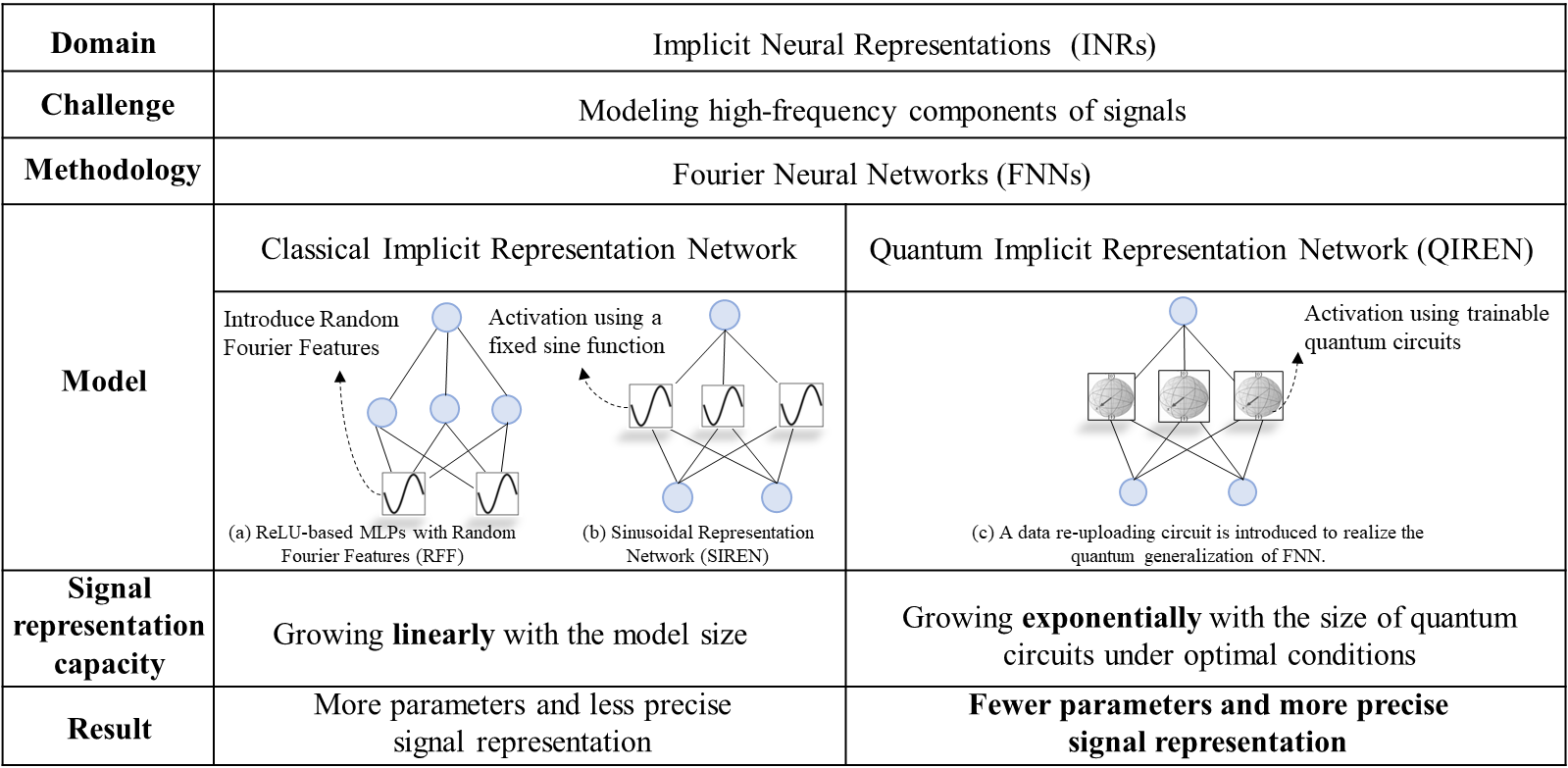}
\caption{Classical Fourier Neural Networks vs. Quantum Fourier Neural Networks.}
\label{fig21}
\end{figure*}
On the other hand, quantum machine learning has recently been given heavy prominence in accelerating computing and saving parameters. Among them, Quantum Neural Networks \cite{farhi2018classification}, also known as variational circuits \cite{mcclean2016theory}, or quantum circuit learning \cite{mitarai2018quantum}, have demonstrated quantum advantages over classical methods in various domains, including classification \cite{schuld2020circuit}, generative adversarial learning \cite{dallaire2018quantum}, and deep reinforcement learning \cite{lockwood2020reinforcement}. However, there is relatively little work demonstrating quantum advantages in real-world tasks that outperform classical methods. Identifying practical applications that reveal the benefits of quantum machine learning has been a goal of this community. Recent advances \cite{schuld2021effect,yu2022power} have found that data re-uploading quantum circuits, which involve uploading data repeatedly into quantum circuits, can be equivalently represented as Fourier series in the data. This inspires us to explore the application of Quantum Neural Networks with data re-uploading quantum circuits as core components in the field of INRs. We hope to achieve a more fine-grained representation of the signal with fewer parameters, as well as to find suitable task scenarios for quantum machine learning.

In this paper, we first demonstrate the exponential advantage of the data re-uploading quantum circuit over classical methods in representing Fourier series under optimal conditions. By combining these quantum circuits with classical layers, we have designed QIREN. Theoretically, QIREN has a stronger ability to fit the Fourier series than classical
FNNs. This is reflected in the fact that QIREN can achieve a more refined signal representation with fewer parameters. Experimental results verify that QIREN indeed exhibits excellent performance, with a reduction in fitting error of up to 35$\%$ and a smaller number of parameters, compared to the SOTA model on the signal representation task. Additionally, it demonstrates improvements in performance for both image superresolution and image generation tasks. We exhibit the core ideas and main conclusions of this paper in Figure \ref{fig21}.

In summary, the following novel contributions are made in this paper:
\begin{itemize}
\item We propose QIREN, a novel quantum generalization of FNNs and highlight implicit neural representations as a field where the potential advantages of Quantum Neural Networks can truly be harnessed.

\item We analyze the role of quantum circuits and classical layers in QIREN and demonstrate the quantum advantages of QIREN over classical FNNs.

\item Through image representation and sound representation tasks, we show that QIREN outperforms SOTA models in signal representation, particularly in modeling high-frequency signals. Moreover, we have explored the application prospect of QIREN in tasks such as image superresolution and image generation.
\end{itemize}
The following sections are organized as follows. In Section \ref{2}, we review related work on implicit neural representations and data re-uploading quantum circuits. In Section \ref{3}, we introduce classical implicit representation networks. In Section \ref{4}, we propose quantum implicit representation networks and theoretically demonstrate the quantum advantage over classical implicit representation networks. In Section \ref{5}, we conduct experiments and show that QIREN has performance beyond the SOTA model in the field of INRs. We ﬁnally present our conclusions and outlook for future work in Section \ref{6}.



\section{Related Work}\label{2}
\subsection{Implicit Neural Representations}
The original concept of INRs was first introduced in Compositional Pattern Producing Network \cite{stanley2007compositional}, which is a neuroevolution-based model that is trained to represent 2D images. Subsequently, INRs gained significant popularity in the field of 3D computer vision due to its provision of continuous, memory-efficient implicit representations for shape parts \cite{genova2019learning,zhong2019reconstructing}, objects \cite{park2019deepsdf}, or scenes \cite{sitzmann2019scene,jiang2020local}. However, early works mostly utilized ReLU-based MLPs, which lacked accuracy in modeling the high-frequency components of signals \cite{xu2019frequency,rahaman2019spectral,tancik2020fourier}. \citet{mildenhall2021nerf} significantly improved the capability of modeling high-frequency details in ReLU-based MLPs by applying Random Fourier Features (RFF) mapping to the inputs. \citet{sitzmann2020implicit} proposed Sinusoidal Representation Network (SIREN), which effectively represents high-frequency signals by using the sine activation function. The models proposed in these two works are considered SOTA models in the field of INRs and have inspired many subsequent works \cite{woo2023learning,huang2022universal,skorokhodov2021adversarial,chan2021pi}. Nevertheless, the capacity of these classical SOTA models to represent signals is constrained by their linear growth with model size. Our proposed model achieves a stronger signal representation capability, by leveraging the exponential advantage of the data re-uploading quantum circuit over classical methods in representing Fourier series.

\subsection{Data Re-uploading Quantum Circuits}
\citet{perez2020data} first introduced the concept of data re-uploading quantum circuits and demonstrated their capability in handling certain point set classification tasks. \citet{wach2023data} further explored their classification performance and conducted experiments on image classification. Some studies concentrate on investigating the mathematical properties of data re-uploading quantum circuits. \citet{schuld2021effect} revealed the fundamental nature of data re-uploading quantum circuits as Fourier series. \citet{yu2022power} further investigated the universality of single-qubit data re-uploading circuits. These works only explore the multi-layer single-qubit circuits or single-layer multi-qubit circuits. We extended our research to multi-layer multi-qubit quantum circuits. \citet{shin2023exponential} proposed an encoding scheme that enables exponential growth of the spectrum of the data re-uploading circuit. On this basis we consider the coefficients of the data re-uploading circuit and analyze them, thus demonstrating that under optimal conditions data re-uploading circuits have an exponential advantage over classical methods in fitting the Fourier series. Meanwhile, our approach is more practical, unlike the above studies which only perform experiments to verify mathematical properties, we conducted experiments on real-world datasets and establish a bridge between data re-uploading circuits and implicit neural representations.

\section{Classical Implicit Representation Network}\label{3}
Two SOTA models in the field of implicit neural representations, ReLU-based MLPs with Random Fourier Features and Sinusoidal Representation Network, have been shown to be FNNs by \citet{benbarka2022seeing}. We review this in this section and further demonstrate that the signal representation capability of these classical FNNs grows linearly with model size.


A perceptron with RFF has the following form:
\begin{equation}\label{eq5}
g(\mathbf{x})=\mathbf{W} \cdot \left(\begin{array}{c}
\cos (2 \pi \mathbf{M} \cdot \mathbf{x}) \\
\sin (2 \pi \mathbf{M} \cdot \mathbf{x})
\end{array}\right) +b,
\end{equation}
where $\mathbf{W} \in \mathbb{R}^{1 \times 2m}$ is the parameter matrix, $\mathbf{M} \in \mathbb{R}^{m \times d_{in}}$ is the random Fourier mapping
matrix ($m$ determines the size of the matrix, serving as a measure of the model size), $\mathbf{x} \in \mathbb{R}^{d_{in}}$ is the input and $b \in \mathbb{R}$ is the bias. Next, we will relate Eq. (\ref{eq5}) to the general equation of Fourier series.

A Fourier series is a weighted sum of sine and cosine functions. When the number of terms approaches infinity, the Fourier series can approximate any periodic function. Even though the function $g(\mathbf{x})$ may not be periodic, the inputs are generally bounded. Therefore, we can assume that $g(\mathbf{x})$ is periodic outside the boundaries of its inputs and its Fourier series takes the following form: 
\begin{equation}\label{eq1}
g(\mathbf{x})=\sum_{\mathbf{n} \in \mathbb{Z}^{d_{in}} } c_{\mathbf{n}} \mathrm{e}^{ 2 \pi \mathrm{i} \frac{ \mathbf{n}}{\mathbf{T}}\cdot \mathbf{x}},
\end{equation}
where $c_{\mathbf{n}}$ denotes the coefficient, and ${\mathbf{T}}$ denotes the period of $g(\mathbf{x})$. For real-valued functions, it holds that $c_{\mathbf{n}} = c_{\mathbf{-n}}^*$. Therefore Eq. (\ref{eq1}) can be rewritten as
\begin{equation}\label{eq2}
g(\mathbf{x})=\sum_{\mathbf{n} \in \mathbb{Z}^{d_{in}}} a_{\mathbf{n}} \cos (2 \pi \frac{\mathbf{n}}{\mathbf{T}}\cdot \mathbf{x})+b_{\mathbf{n}} \sin (2 \pi \frac{ \mathbf{n}}{\mathbf{T}}\cdot \mathbf{x}).
\end{equation}
If we write Eq. (\ref{eq2}) in vector form, we get
\begin{equation}\label{eq3}
g(\mathbf{x})=\left(a_{\mathbf{N}}, b_{\mathbf{N}}\right) \cdot\left(\begin{array}{c}
\cos (2 \pi \mathbf{N} \cdot \mathbf{x}) \\
\sin (2 \pi \mathbf{N} \cdot \mathbf{x})
\end{array}\right),
\end{equation}
where $\mathbf{N}= \{\frac{\mathbf{n}}{\mathbf{T}}\}_{\mathbf{n} \in  \mathbb{Z}^{d_{in}}}$ denotes frequency spectrum. Now, when we compare Eq. (\ref{eq5}) and (\ref{eq3}), we find similarities. Obviously, a perceptron with RFF constructs a Fourier series, where $\mathbf{W}$ includes coefficients of $g(\mathbf{x})$, and $\mathbf{M}$ includes frequencies of $g(\mathbf{x})$.

\begin{figure*}[h]
\centering
\includegraphics[scale=0.40]{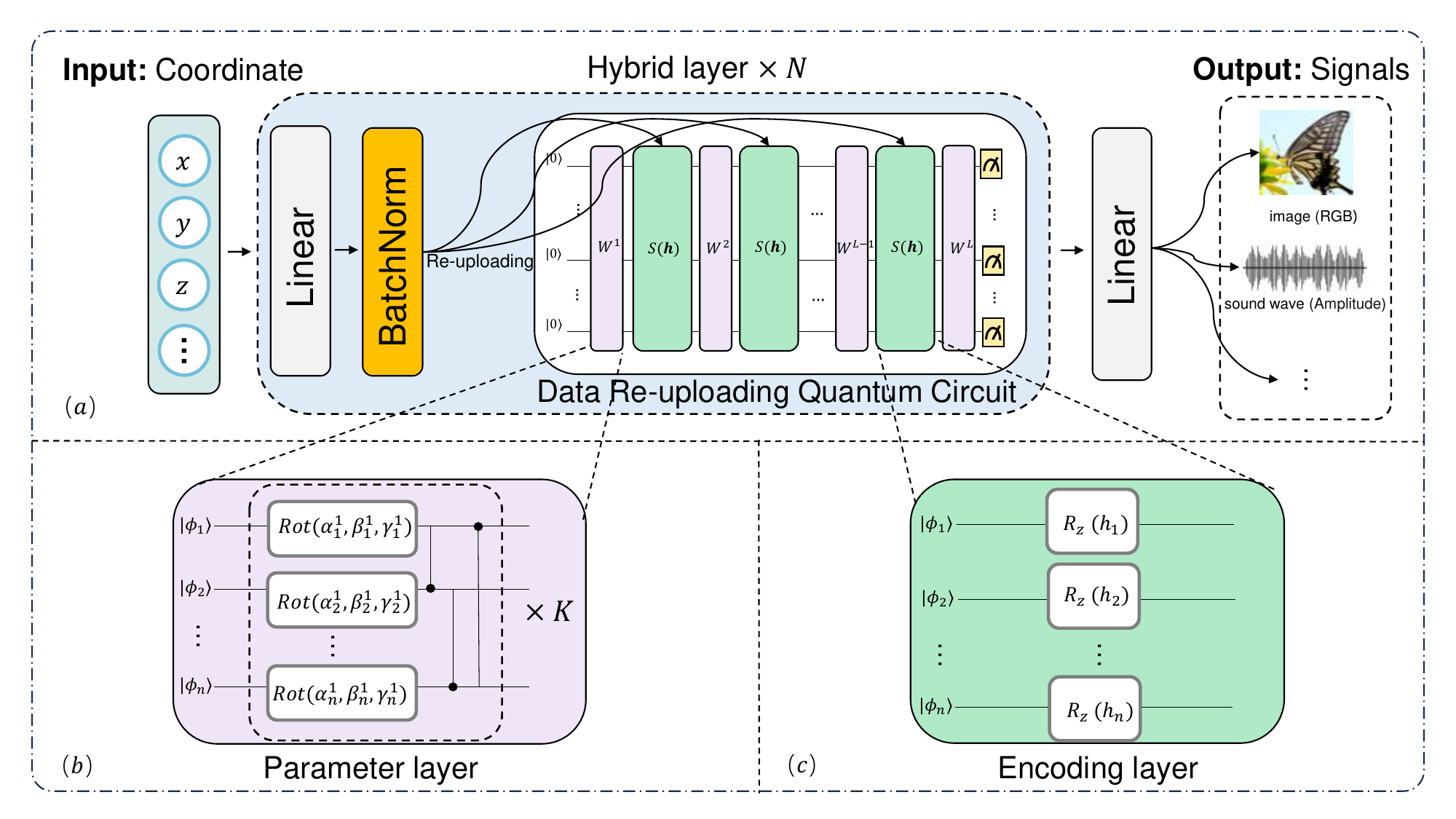}
\caption{Architecture of QIREN. (a) presents the overall architecture of QIREN. (b) and (c) respectively illustrate the implementation details of the parameter layer and the encoding layer.}
\label{fig2}
\end{figure*}

SIREN is a MLP that utilizes the sine activation function. Considering SIREN with one hidden layer, we get
\begin{equation}\label{eq6}
g(\mathbf{x})=\mathbf{W_2} \cdot \sin (2 \pi \mathbf{W_1} \cdot \mathbf{x}+\boldsymbol{\theta})+b,
\end{equation}
where $\mathbf{W_1} \in \mathbb{R}^{2m \times d_{in}}$ and $\mathbf{W_2} \in \mathbb{R}^{1 \times 2m}$ are the parameter matrices and $\boldsymbol{ \theta } \in \mathbb{R}^{2 m}$ is the bias vector. We can consider a specific case when $\boldsymbol{ \theta }=(\pi / 2, \ldots, \pi / 2,0, \ldots, 0)^{T}$. In this case, the SIREN and the perceptron with RFF have nearly identical forms, except that $\mathbf{W_2}$ is a trainable parameter matrix and $\mathbf{M}$ is an untrained random matrix. In fact, any situation can be transformed into this special case, as we can always express Eq. (\ref{eq6}) as
\begin{equation}
g(\mathbf{x})=\mathbf{W_2} \cdot \sin (2 \pi \mathbf{W_1} \cdot \mathbf{x}+ \boldsymbol{\hat{\theta} } +\left(\begin{array}{c}
\boldsymbol{\frac{\pi}{2}}  \\
\mathbf{0}
\end{array}\right) )+b,
\end{equation}
where $\boldsymbol{\hat{\theta}}=\boldsymbol{\theta} - \left(\begin{array}{c}
\boldsymbol{\frac{\pi}{2}}  \\
\mathbf{0}
\end{array}\right).$

At this point, we can conclude that both ReLU-based MLP with RFF and SIREN are FNNs as their fundamental blocks approximate functions in the form of Fourier series. Their performance is directly correlated with the size of the Fourier series they can represent. \textbf{Therefore, the signal representation capacity of classical FNNs increases linearly with the model size $m$}.

\section{Quantum Implicit Representation Network}\label{4}
In Section \ref{4.1} and \ref{4.2}, we introduce the architecture of QIREN and its nature as a Fourier Neural Network. In Section \ref{4.3}, we analyze the role of each component in QIREN.

\subsection{Overall Architecture of QIREN}\label{4.1}
The overall architecture of QIREN is shown in Figure \ref{fig2}(a), which consists of $N$ hybrid layers and a Linear layer at the end. The model takes coordinates as inputs and outputs signal values. Correspondingly, the dataset $D = \{(\mathbf{x}_i ,\mathbf{y}_i)\}$ is a set of tuples of coordinates $\mathbf{x}_i \in \mathbb{R}^{d_{in}}$ along with signals $\mathbf{y}_i \in \mathbb{R}^{d_{out}}$. Next, we provide a detailed description of the model.

The data $\mathbf{x}_i$ initially enters the hybrid layer, starting with a Linear layer and a BatchNorm layer. We get
\begin{equation}\label{eq7}
\mathbf{h}_i = \text{BatchNorm}(\mathbf{W}\mathbf{x}_i + \mathbf{b}),
\end{equation}
where $\mathbf{W}$ and $\mathbf{b}$ are parameters of the Linear layer. Then $\mathbf{h}_i$ is input to the data re-uploading quantum circuit $\text{QC}$ (Preliminaries of the quantum circuit is presented in the Appendix \ref{qc}). In Figure \ref{fig2}(b) and (c), we present the implementation structures for the parameter layer $W$ and encoding layer $S(\mathbf{h}_i)$ of the quantum circuit. The parameter layer consists of $K$ stacked blocks. Each block contains $Rot$ gates applied to each qubit and $CNOT$ gates that connects qubits in a cyclic manner. The encoding layer applies $R_Z$ gates on each qubit. Finally, we measure the expectation value of the quantum state with observables. The output of the quantum circuit is given by
\begin{equation}\label{eq8}
\begin{array}{r}
\mathbf{f}_{i}=[\text{QC}(\mathbf{h}_i,O_1),\ldots,\text{QC}(\mathbf{h}_i,O_{d_f})],
\end{array}
\end{equation}
where $O$ denotes an arbitrary observable. The output $\mathbf{f}_{i}^{(n)}$ of the $n$-th hybrid layer will be used as the input of the $(n+1)$-th layer. At the end, we add a Linear layer to receive $\mathbf{f}_{i}^{(N)}$ and output $\mathbf{y'}_i$. We use Mean Squared Error (MSE) as the loss function to train the model:
\begin{equation}\label{eq10}
\text{MSE} =\frac{1}{\left |D  \right | } \sum_{i}\left(\mathbf{y}_i-\mathbf{y'}_i\right)^{2}.
\end{equation}


\subsection{QIREN as a Quantum Generalization of FNNs}\label{4.2}
\citet{schuld2021effect} derives that the data re-uploading circuit is essentially the Fourier series and analyzes it on single-layer multi-qubit and multi-layer single-qubit circuits. Based on this, we generalize the analysis to multi-layer multi-qubit circuits and show that QIREN is essentially a Fourier Neural Network.

Each component of Eq. (\ref{eq8}) can be defined as $f(\mathbf{h})$, given by the multi-layer multi-qubit data re-uploading quantum circuit, which is represented as follows:
\begin{equation}\label{eq11}
f(\mathbf{h}) =\left\langle 0\left|U^{\dagger}(\mathbf{h}) O U(\mathbf{h})\right| 0\right\rangle,
\end{equation}
where
\begin{equation}\label{eq9}
U(\mathbf{h})=W^{(L)} S(\mathbf{h})  W^{(L-1)} \ldots W^{(2)} S(\mathbf{h}) W^{(1)}.
\end{equation}
In the data re-uploading quantum circuit, each encoding layer has the form $S(\mathbf{h}):=\mathrm{e}^{-\mathrm{i} h_{1} H} \otimes \ldots \otimes \mathrm{e}^{-\mathrm{i} h_{d_{h}} H}$, where $H$ is a $d$-qubit Hamiltonian that depends on the actual quantum gates used, and there always exists an eigenvalue decomposition $H=V^{\dagger} \Sigma V$. Without loss of generality, we can assume that all Hamiltonians are diagonal as $V$ and $V^{\dagger}$ can be absorbed into adjacent parameter layers. With this assumption, we note that the $S(\mathbf{h})$ is diagonal and $l$-th $S(\mathbf{h})$ has the form:
\begin{equation}\label{eq12}
{[S(\mathbf{h})]_{(j_{1}^{(l)}, \ldots, j_{d_{h}}^{(l)}), (j_{1}^{(l)}, \ldots, j_{d_{h}}^{(l)})}=\mathrm{e}^{-\mathrm{i} \left(\lambda_{j_{1}^{(l)}}, \ldots, \lambda_{j_{d_{h}}^{(l)}}\right) \cdot \mathbf{h} }}, 
\end{equation}
where $\lambda$ is the eigenvalue of $H$. We can rewrite Eq. (\ref{eq12}) by introducing the multi-index $\mathbf{j}^{(l)}=\left\{j_{1}^{(l)}, \ldots, j_{d_{h}}^{(l)}\right\}\in \{1, \ldots, 2^{d}\}^{d_{h}}$ as follows: 
\begin{equation}\label{eq13}
{[S(\mathbf{h})]_{\mathbf{j}^{(l)} , \mathbf{j}^{(l)}}=\mathrm{e}^{-\mathrm{i} \boldsymbol{\lambda}_{\mathbf{j}^{(l)}} \cdot \mathbf{h} }}, 
\end{equation}
where $\boldsymbol{\lambda}_{\mathbf{j}^{(l)}}=\left(\lambda_{j_{1}^{(l)}}, \ldots, \lambda_{j_{d_{h}}^{(l)}}\right)$. In this way, we can derive the mathematical expression of the quantum state $|\psi \rangle$:
\begin{equation}\label{eq14}
\begin{array}{r}
{|\psi \rangle_{i}=[U(\mathbf{h})|0\rangle]_{i}=\sum\limits_{\mathbf{j}^{(1)} \ldots \mathbf{j}^{(L-1)}} \mathrm{e}^{-\mathrm{i}\left(\boldsymbol{\lambda}_{\mathbf{j}^{(1)}}+\cdots+ \boldsymbol{\lambda}_{\mathbf{j}^{(L-1)}} \right) \cdot \mathbf{h}} } \\
 \times W_{i \mathbf{j}^{(L-1)}}^{(L)} \ldots W_{\mathbf{j}^{(2)} \mathbf{j}^{(1)}}^{(2)} W_{\mathbf{j}^{(1)} 1}^{(1)},
\end{array}
\end{equation}
where the subscript of $W$ indicates the row and column. We introduce another multi-index $\mathbf{J}=\left\{\mathbf{j}^{(1)}, \ldots, \mathbf{j}^{(L-1)}\right\} \in \{1, \ldots, 2^{d}\}^{d_{h}\times(L-1)}$ to simplify Eq. (\ref{eq14}) as follows:
\begin{equation}\label{eq15}
|\psi \rangle_{i}=\sum_{\mathbf{J}} \mathrm{e}^{-\mathrm{i} \boldsymbol{\Lambda}_{\mathbf{J}} \cdot \mathbf{h}}  W_{i \mathbf{j}^{(L-1)}}^{(L)} \ldots W_{\mathbf{j}^{(2)} \mathbf{j}^{(1)}}^{(2)} W_{\mathbf{j}^{(1)} 1}^{(1)},
\end{equation}
where $\boldsymbol{\Lambda}_{\mathbf{J}}=\boldsymbol{\lambda}_{\mathbf{j}^{(1)}}+\cdots+\boldsymbol{\lambda}_{\mathbf{j}^{(L-1)}}$.
Next, according to Eq. (\ref{eq11}), we calculate the expectation value of $|\psi\rangle$ and its conjugate transpose $\langle\psi|$ with respect to $O$, and get
\begin{equation}\label{eq16}
\begin{array}{r}
f(\mathbf{h}) =\langle\psi|O|\psi\rangle=\sum\limits_{\mathbf{K}, \mathbf{J}} \mathrm{e}^{\mathrm{i}\left(\boldsymbol{\Lambda}_{\mathbf{K}}-\boldsymbol{\Lambda}_{\mathbf{J}}\right) \cdot \mathbf{h}} \sum\limits_{i, i^{\prime}}{W^{\dagger }}_{\mathbf{k}^{(1)} 1}^{(1)}\\{W^{\dagger }}_{\mathbf{k}^{(2)}\mathbf{k}^{(1)}}^{(2)}\ldots{W^{\dagger }}_{i \mathbf{k}^{(L-1)}}^{(L)} O_{i, i^{\prime}}  W_{i^{\prime} \mathbf{j}^{(L-1)}}^{(L)} \ldots W_{\mathbf{j}^{(2)} \mathbf{j}^{(1)}}^{(2)} W_{\mathbf{j}^{(1)} 1}^{(1)}.
\end{array}
\end{equation}
We are currently not concerned with the specific form of the observable $O$ and the parameter layers. So we substitute $a_{\mathbf{K}, \mathbf{J}}$ for the second summation term in Eq. (\ref{eq16}), resulting in a more intuitive form:
\begin{equation}\label{eq17}
f(\mathbf{h})=\sum_{\mathbf{K}, \mathbf{J}} a_{\mathbf{K}, \mathbf{J}} \mathrm{e}^{\mathrm{i}\left(\boldsymbol{\Lambda}_{\mathbf{K}}-\boldsymbol{\Lambda}_{\mathbf{J}}\right) \cdot \mathbf{h}},
\end{equation}
where $a_{\mathbf{K}, \mathbf{J}}$ denotes the coefficient, $\boldsymbol{\Lambda}_{\mathbf{K}}-\boldsymbol{\Lambda}_{\mathbf{J}}$ denotes the frequency and $\{\boldsymbol{\Lambda}_{\mathbf{K}}-\boldsymbol{\Lambda}_{\mathbf{J}}\}_{\mathbf{K}, \mathbf{J}}$ denotes the frequency spectrum. Eq. (\ref{eq17}) is now evidently equivalent to the Fourier series. Utilizing this quantum circuit enables QIREN to approximate functions in the form of the Fourier series. Thus, we have demonstrated that QIREN is a quantum generalization of FNNs.

\begin{table*}[h] 
\setlength{\tabcolsep}{2.0pt}
\small
\centering
\begin{tabular}{c|ccc|ccccc|ccc} 
\toprule 
\multirow{2}{*}{Method} &\multicolumn{3}{c}{Sound Representation}&\multicolumn{5}{|c|}{Image Representation}&\multicolumn{3}{c}{Image Superresolution}\\  &Cello&$\#$params&$\#$mem($\%$)&Astronaut&Camera&Coffee&$\#$params&$\#$mem($\%$)&Astronaut&Camera&Coffee\\ 
\midrule
Nearest &-&-&-&-&-&-&-&-&26.6&10.4&13.6\\
Bilinear &-&-&-&-&-&-&-&-&25.2&9.2&12.2\\
ReLU&6.8&831&16.9&9.9&2.7&4.2&841&17.9&33.8&11.3&13.5\\
Tanh&14.0&831&16.9&20.7&5.8&14.8&841&17.9&47.8&15.2&26.7\\
$\text{ReLU+RFF}^\star$ &6.0&791&20.9&5.1&1.9&4.9&791&22.8&39.9&13.3&23.9\\ 
$\text{SIREN}^\star$ &\textbf{5.5}&691&30.9&9.0&1.5&2.3&701&31.5&77.0&26.3&15.7\\
\textbf{QIREN} (ours)&\textbf{5.5}&\textbf{649}&\textbf{35.1}&\textbf{4.0}&\textbf{1.1}&\textbf{1.5}&\textbf{657}&\textbf{35.8}&\textbf{24.3}&\textbf{7.9}&\textbf{9.4}\\ 
\bottomrule 
\end{tabular}
\caption{MSE ($\times 10^{-3}$) of different models on signal representation and image superresolution tasks. The best results are highlighted in \textbf{bold}. The models widely regarded as SOTA are marked with $\star$. $\#$params denote the number of model parameters, and $\#$mem denotes the memory saved by the model compared to the discrete grid-sampled representation. }
\label{tb1}
\end{table*}

\subsection{Theoretical Analysis of QIREN}\label{4.3}
In this subsection, we analyze the roles of each component in QIREN and demonstrate the quantum advantages of QIREN over classical FNNs. We summarize them into three claims.~\\

\noindent \textbf{Claim 1} \textit{The capability of the data re-uploading quantum circuit to represent Fourier series grows exponentially with the size of the circuit under optimal conditions.}~\\

We consider a specific quantum circuit with $d \times d_{h}$ qubits, where the data is repeatedly uploaded $L$ times, and $H=\sum_{q=1}^{d} Z^{(q)}/2$ ($Z^{(q)}$ denotes Pauli $Z$ acting on the $q$-th qubit). Regarding $H$, it has $2^d$ eigenvalues with the $d+1$ unique entries:
\begin{equation}\label{eq18}
\mathbf{P}=\{p-\frac{d}{2}, \mid p \in\{0, \ldots, d\}\}.
\end{equation}
In this case, the frequency spectrum of $f(\mathbf{h})$ can be represented as
\begin{equation}\label{eq19}
\begin{aligned}
\{\boldsymbol{\Lambda}_{\mathbf{K}}-&\boldsymbol{\Lambda}_{\mathbf{J}}\}_{\mathbf{K}, \mathbf{J}} \\ &=\left\{\boldsymbol{\lambda}_{\mathbf{k}^{(1)}}+\cdots+\boldsymbol{\lambda}_{\mathbf{k}^{(L)}}-(\boldsymbol{\lambda}_{\mathbf{j}^{(1)}}+\cdots+\boldsymbol{\lambda}_{\mathbf{j}^{(L)}})\right\} \\
& =\left\{\left(\lambda_{k_{1}^{(1)}}+\cdots+\lambda_{k_{1}^{(L)}}\right)-\left(\lambda_{j_{1}^{(1)}}+\cdots+\lambda_{j_{1}^{(L)}}\right) \mid\right. \\
&\hspace{2em} \left.\lambda_{k_{1}^{(1)}}, \ldots, \lambda_{k_{1}^{(L)}}, \lambda_{j_{1}^{(1)}}, \ldots,\lambda_{j_{1}^{(L)}} \in \mathbf{P} \right\}^{d_{h}} \\
&  =\{-dL, \ldots, dL\}^{d_{h}}.
\end{aligned}
\end{equation}
It can be seen that the spectrum grows exponentially with $d_h$. \textbf{With the help of the Linear layer, the spectrum can grow exponentially with the size of the circuit $d \times d_{h}$ at most, as will be explained in Claim 2.} However, one unsatisfactory point is that we cannot guarantee that the coefficients $a_{\mathbf{K}, \mathbf{J}}$ are arbitrary because this would require us to have the ability to construct arbitrary unitary matrices to some extent. According to the Solovay-Kitaev theorem \cite{nielsen2010quantum} which follows that any arbitrary unitary matrix on $n$ qubits can be approximated within an error $\epsilon$ using $O\left(n^{2} 4^{n} \log ^{c}\left(n^{2} 4^{n} / \epsilon\right)\right)$ two-qubit gates ($c$ is a constant), constructing arbitrary unitary matrices would consume exponential resources.

Fortunately, the coefficients of most functions we need to fit in practice are not arbitrary but rather follow certain prior distributions (e.g., the amplitudes of high-frequency components are often small in natural signals). This can draw an analogy to quantum many-body physics, where the majority of natural systems have Hamiltonians that can be expressed as a sum of local Hamiltonians \cite{gao2022enhancing}. Theorem 1 \cite{poulin2011quantum} states that approximating the evolution process of such a system only requires a polynomial number of parameters.

\noindent \textbf{Theorem 1} \textit{The time evolution operator corresponding to a Hamiltonian composed of $L$ $k$-body terms, with a total evolution time of $T$,  can be simulated by a quantum circuit of polynomial size. The total number of standard universal 2-qubit quantum gates needed to approximate the complete time evolution with an error $\epsilon$ is upper bounded by}
\begin{equation}
G_{\text {tot }}=d_{S K} \frac{c_{\max }^{2}T^{2} L^{3}}{\epsilon}  \log ^{c S K}(\frac{c_{\max }^{2}T^{2} L^{3}}{\epsilon^2})
\end{equation}
\textit{where $cSK$ and $dSK$ are constants, $c_{\max }$ is the maximum norm of the k-body terms in the Hamiltonian.}

\begin{figure*}[h]
\centering
\includegraphics[scale=0.65]{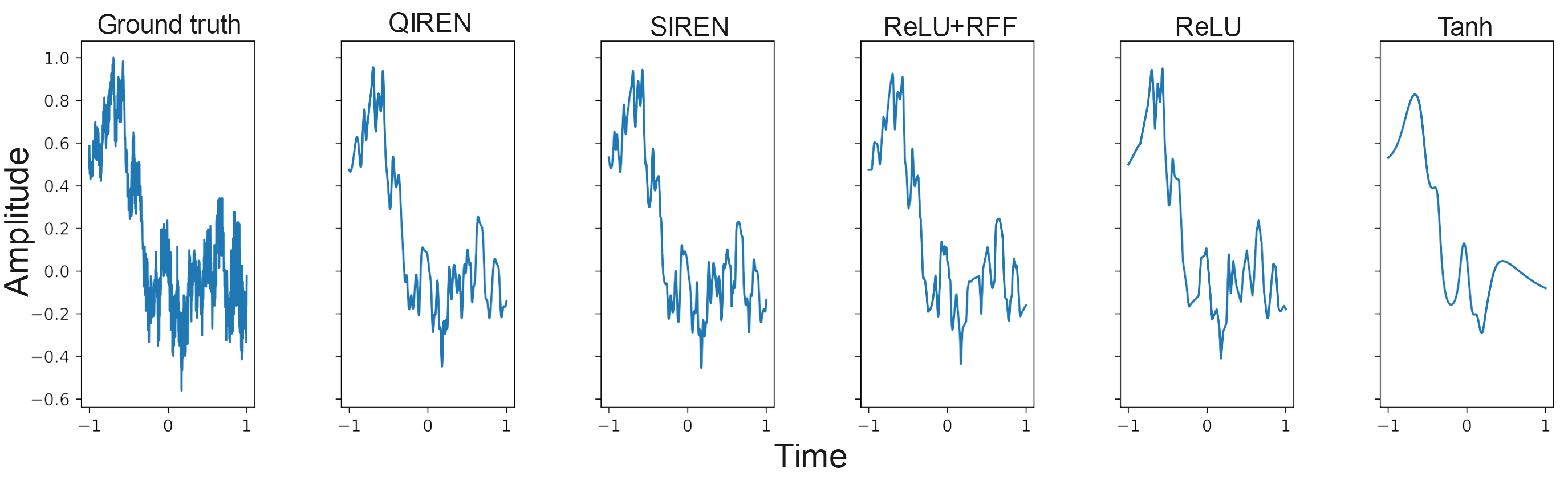}
\caption{Results of sound representation. The Y-axis represents the normalized amplitude of the sound wave. The X-axis represents the time.}
\label{fig3}
\end{figure*}

\begin{figure*}[h]
\centering
\includegraphics[scale=0.49]{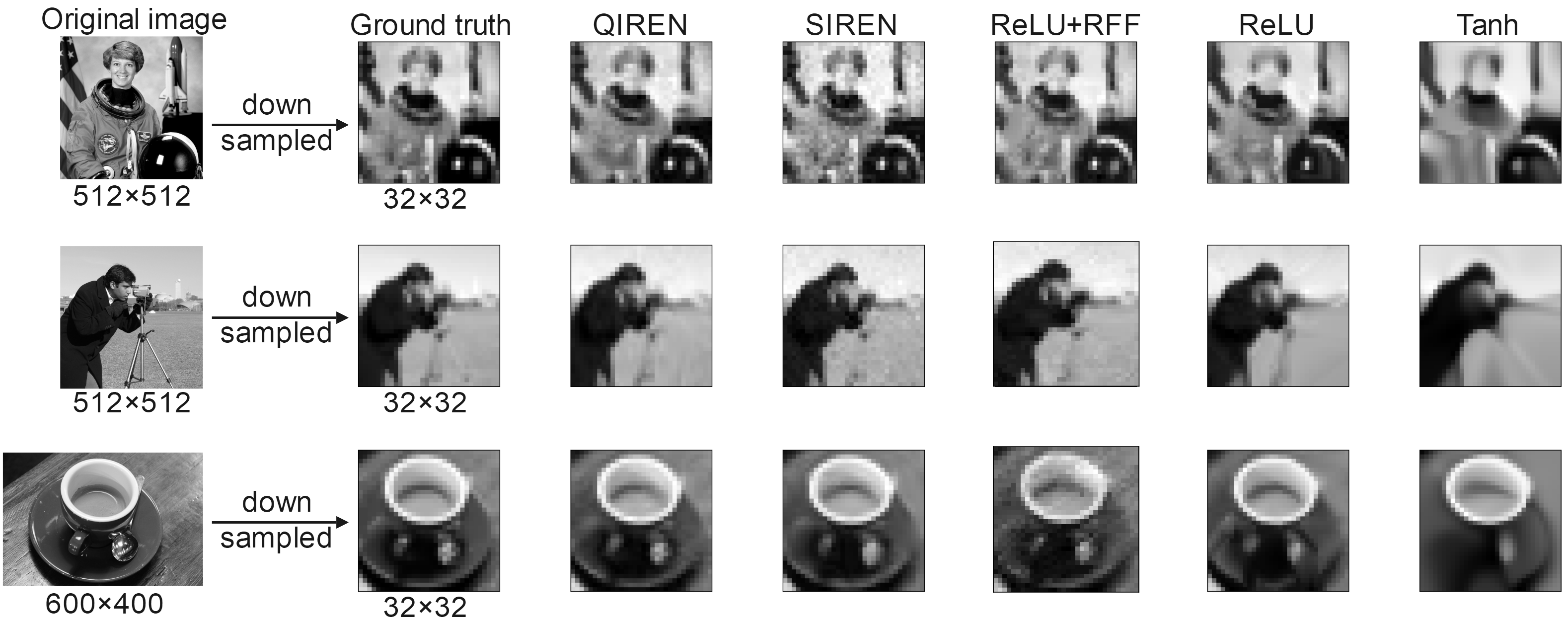}
\caption{Results of image representation.}
\label{fig4}
\end{figure*}

Hence, we can assume that the Hamiltonian corresponding to the function we aim to fit can also be decomposed into a sum of local Hamiltonians. When the ansatz of the parameter layer is well-adapted to approximate this Hamiltonian, a polynomial number of parameters in the parameter layer is sufficient to effectively fit the function. At this point, we can conclude Claim 1. Furthermore, utilizing this quantum circuit can empower QIREN with quantum advantages over classical FNNs.


~\\
\noindent \textbf{Claim 2} \textit{The role of the Linear layer is to further expand the spectrum and adjust the frequency, leading to improved fitting performance.}~\\

By observing Eq. (\ref{eq13}), we can understand that applying a Linear layer before uploading the data to the quantum circuit is equivalent to adjusting the eigenvalues in the Hamiltonian, ultimately affecting the frequency spectrum. Using the Linear layer brings two advantages. First, it can make the frequency spectrum larger. From Eq. (\ref{eq18}) and (\ref{eq19}), it can be deduced that certain sums of eigenvalues generate the same term in the spectrum. This redundancy can be reduced by using the Linear layer, and the size of the spectrum can be extended from $(2dL+1)^{d_h}$ to up to $((3^d-1)L+1)^{d_h}$. Second, it enables the frequency spectrum to be adjustable, aiming to cover frequencies with larger coefficients. Therefore, incorporating Linear layers can further enhance the fitting performance of QIREN. A more detailed explanation of Claim 2 can be found in Appendix \ref{claim2}. ~\\

\noindent \textbf{Claim 3} \textit{The role of the Batchnorm layer is to accelerate the convergence of our quantum model.}~\\

In feedforward neural networks, the data often passes through a BatchNorm layer before the activation function, which effectively prevents the vanishing gradient problem \cite{ioffe2015batch}. Similarly, in QIREN, the quantum circuits replace activation functions and play a role in providing non-linearity (The quantum circuit itself is linear, but the process of uploading classical data to the quantum circuit is non-linear). Therefore, we add the BatchNorm layer here, with the aim of stabilizing and accelerating the convergence of the model. We have further validated its effectiveness through ablation experiments in Appendix \ref{aa}.

\section{Experiments}\label{5}
In Section \ref{5.1}, we validate the superior performance of QIREN in representing signals, especially high-frequency signals, through image representation and sound representation tasks. In Section \ref{5.2} and \ref{5.3}, we show the expanded applications of QIREN in tasks such as image superresolution and image generation. To ensure a fair comparison, we adjust the hyperparameters to approximately match the number of parameters between the QIREN and the baseline models. In each experiment, each model is trained five times, and the best performance is selected as the comparison result. Further details regarding the model implementation can be found in Appendix \ref{imp}. All our experiments are conducted on a simulation platform, utilizing the Pennylane \cite{bergholm2018pennylane} and TorchQauntum \cite{wang2022quantumnas}.

\begin{figure}[t]
\centering
\includegraphics[scale=0.35=6]{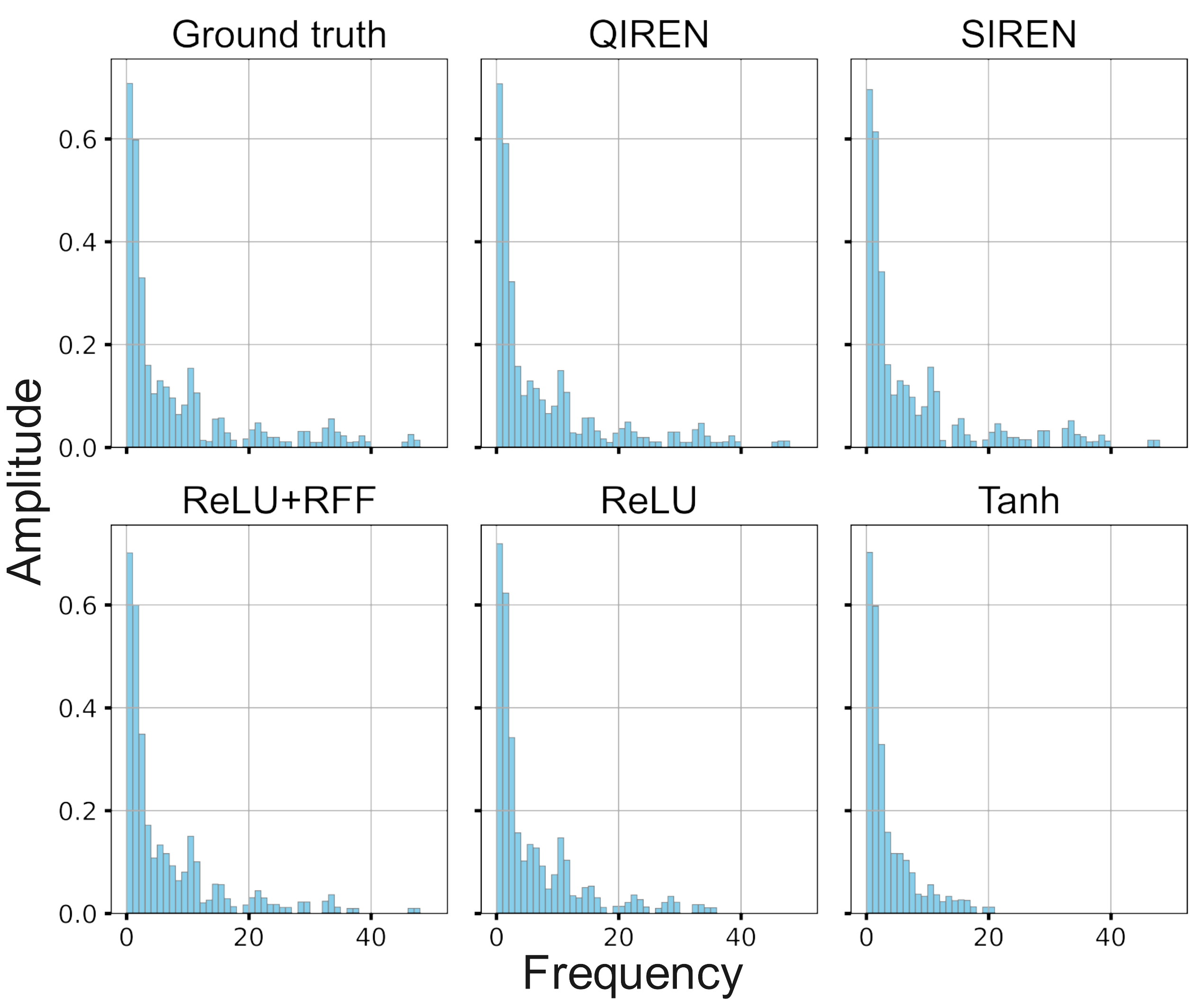}
\caption{The frequency spectrum of the output of models on the sound representation task.}
\label{fig5}
\end{figure}

\subsection{Signal Representation}\label{5.1}
\textbf{Task definition.} The goal is to use neural networks to represent signals. Specifically, we aim to train a model that takes one-dimensional or two-dimensional coordinates as input and outputs the corresponding amplitude or grayscale value at that position.

\textbf{Datasets and Evaluation.} In the sound representation task, we take a small section of one movement from Bach's Cello Suites, sample 1000 points at equal intervals, and normalize the amplitudes. We set the corresponding timestamps to be evenly spaced within $[-1, 1]$. In the image representation task, we utilize three popular images: Astronaut, Camera, and Coffee \cite{van2014scikit}. These images are cropped and down-sampled to dimensions of $32 \times 32$ pixels, resulting in three datasets, each consisting of 1024 pixels. We select the following models as our baselines: ReLU-based MLP, Tanh-based MLP, ReLU-based MLP with RFF \cite{mildenhall2021nerf} and SIREN \cite{sitzmann2020implicit}. MSE is used as a metric to evaluate the difference between the images or sound fitted by the model and the ground truth.

\textbf{Results.} The experimental results are presented in Figure \ref{fig3}, Figure \ref{fig4} and Table \ref{tb1}. QIREN and SIREN exhibit similar performance on the sound representation task. Although the performance of these two models seems comparable, it is worth emphasizing that our model achieves a significant memory saving of $35.1\%$ with the minimum number of parameters and the convergence of SIREN requires proper setting of hyperparameters, whereas our model does not encounter such limitations. Then we explored the output of models from a frequency perspective. We visualize the frequency spectrum of the output of models in Figure \ref{fig5}. It is evident that the low-frequency distribution of the outputs from different models all resembles the ground truth. However, when it comes to the high-frequency distribution, QIREN and SIREN exhibit superior performance, followed by the ReLU-based MLP with RFF. ReLU-based and Tanh-based MLPs even lack the high-frequency parts.

\begin{figure}[t]
\centering
\includegraphics[scale=0.48]{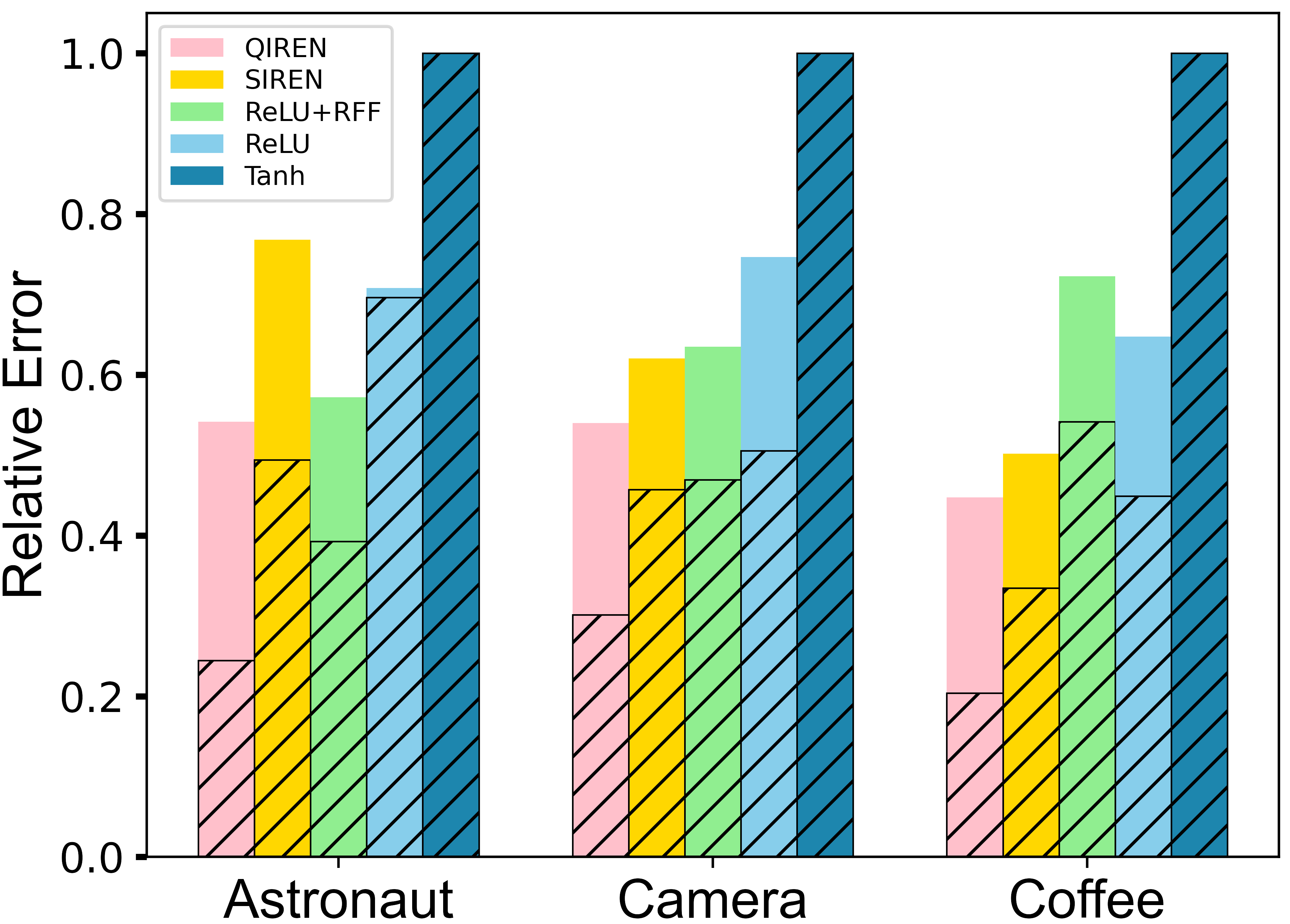}
\caption{The relative error of each model compared to the Tanh-based MLP. The shaded area represents the low-frequency error, while the non-shaded area represents the high-frequency error.}
\label{fig6}
\end{figure}

QIREN achieved optimal performance with the fewest parameters on the image representation task, resulting in a maximum reduction of 34.8$\%$ in error compared to the SOTA model. To further explore the signal representation capabilities of the models, we used filters to separate the high-frequency and low-frequency components of their outputs and compared the fitting errors of these two components individually. The results are presented in the Figure \ref{fig6}. QIREN consistently achieved the lowest error in fitting both high-frequency and low-frequency components.



\subsection{Image Superresolution}\label{5.2}
\textbf{Task definition.} The goal is to increase the resolution of an image while maintaining its content and details as much as possible. In this task, we partition the coordinates into a grid of size $64 \times 64$ and use them as input for the model trained on $32\times 32$ pixel images to reconstruct $64 \times 64$ pixel images.

\textbf{Datasets and Evaluation.} This task directly utilizes the models trained on image representation tasks. The images Astronaut, Camera, and Coffee are down-sampled to create $64 \times 64$ pixel images, which serve as the ground truth for evaluation. For a more comprehensive comparison, in addition to the baseline models used in the signal representation task, we also include bilinear interpolation and nearest-neighbor interpolation. The evaluation metric remains MSE.

\begin{figure*}[h]
\centering
\includegraphics[scale=0.56]{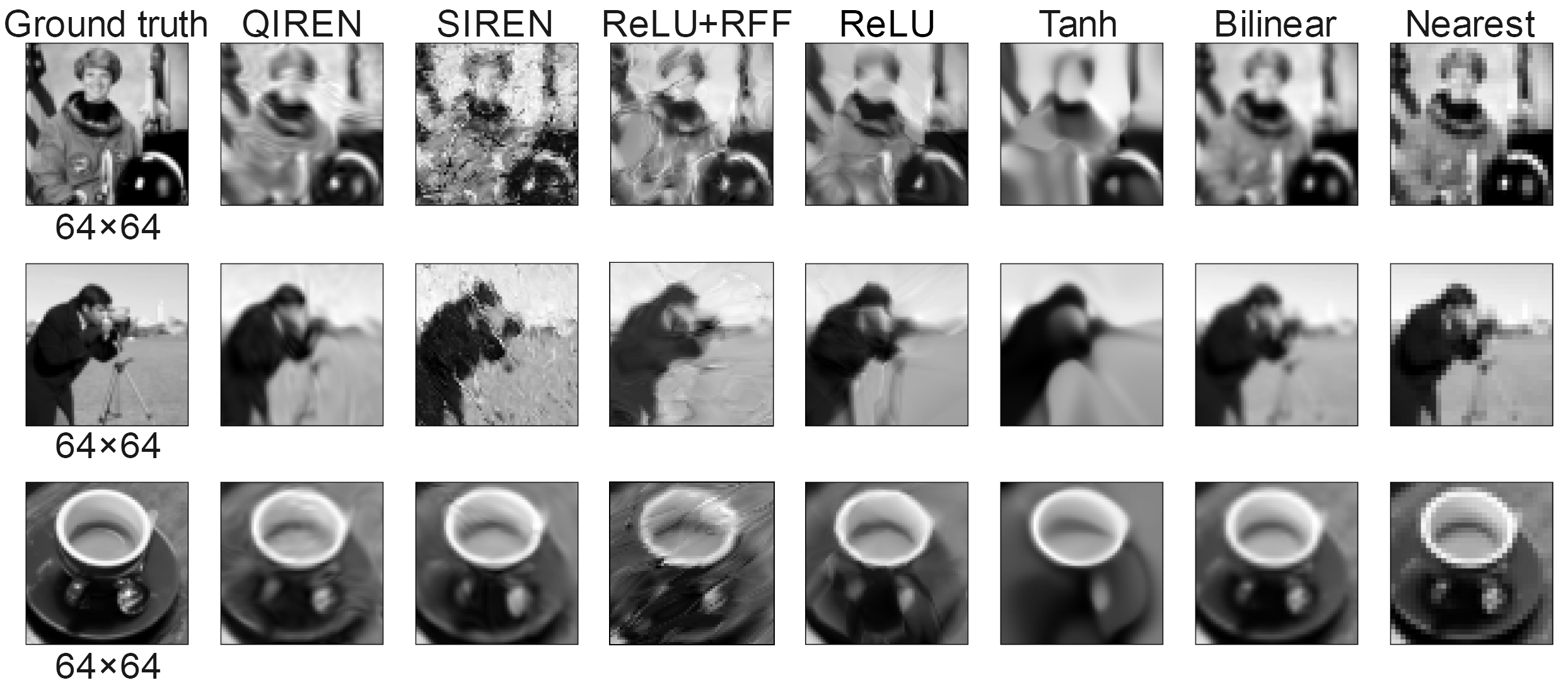}
\caption{Results of image superresolution.}
\label{fig7}
\end{figure*}

\textbf{Results.} The experimental results are presented in Figure \ref{fig7} and Table \ref{tb1}. As observed in Table \ref{tb1}, the performance of classical Implicit Representation Networks failed to surpass that of interpolation methods, possibly due to the limited number of parameters in the network. However, even in such circumstances, QIREN outperforms interpolation methods by achieving a maximum error reduction of 24.0$\%$. This stems from the superior signal representation capability of QIREN, as compared to other models, the parameterized implicit function of QIREN exhibits a distribution that is more similar to the target function.

\subsection{Image Generation}\label{5.3}

\textbf{Task definition.} The goal of this task is to generate facial images that are indistinguishable from real images with the INR-based generator. Recent advancements \cite{skorokhodov2021adversarial,shaham2021spatially,anokhin2021image,dupont2021generative,chan2021pi} 
have introduced a groundbreaking framework and extended INRs to image generation. More specifically, this framework utilizes a hypernetwork \cite{DBLP:journals/corr/HaDL16} that takes random distributions as input to generate the parameters of the Implicit Representation Network. Subsequently, these generated parameters are assigned to the Implicit Representation Network. Finally, the Implicit Representation Network takes coordinate-based inputs and generates images as output. An adversarial approach \cite{goodfellow2020generative} is employed to ensure the generated images align with our desired outcomes. In this task, we adopt such a framework and build upon the foundation of StyleGAN2 \cite{karras2020analyzing}.

\textbf{Datasets and Evaluation.} FFHQ is a high-resolution dataset of 70k human faces \cite{karras2019style} and CelebA-HQ is a high-quality version of CelebA that consists of 30k images \cite{karras2017progressive}. We downsample the images in these two datasets to obtain $32\times 32$ pixel images for experimentation. In this task, we utilize the same baseline models as those used in the signal representation task. We evaluate the model using Frechet Inception Distance (FID) \cite{heusel2017gans} metric using 50k images to compute the statistics.

\textbf{Results.} The experimental results are presented in Table \ref{tb2}. We observed that the ReLU-based MLP is completely inadequate for image generation tasks, whereas RFF can significantly enhance its performance. Tanh-based MLP and SIREN show similar performance. Compared to these classical implicit representation networks, QIREN achieves superior performance with the fewest number of parameters. In Appendix \ref{img_gen}, we will show the images generated by QIREN and further explore its exciting properties, such as out-of-the-box superresolution and meaningful image-space interpolation.

\begin{table}[h] 
\centering
\begin{tabular}{c|ccc} 
\toprule 
Method&FFHQ&CelebA-HQ&$\#$params\\ 
\midrule
Tanh&26.98&25.17&1.16M\\
ReLU&84.94&110.81&1.16M\\
$\text{ReLU+RFF}^\star$&15.01&13.91&1.14M\\ 
$\text{SIREN}^\star$&22.31&20.97&1.16M\\
\textbf{QIREN (ours)}&\textbf{11.53}&\textbf{11.78}&\textbf{1.13M}\\ 
\bottomrule 
\end{tabular}
\caption{FID scores of different models on FFHQ and CelebA-HQ datasets.}
\label{tb2}
\end{table}

\section{Conclusion and Future Work}\label{6}
In this paper, we first derive that QIREN is a quantum generalization of FNNs. Subsequently, we perform a theoretical analysis of the data re-uploading quantum circuits, Linear layers, and BatchNorm layers, demonstrating the ability of QIREN to represent the Fourier series is exponentially stronger over classical FNNs. Lastly, we conduct experiments on signal representation, image superresolution, and image generation. The results demonstrate that QIREN significantly outperforms SOTA models.


It is worth emphasizing that INRs have many other potential applications such as representing scenes or 3D objects, time series forecasting, and solving the differential equation. For a large class of tasks that model continuous signals, we can consider introducing INRs as the basic component. Based on the theoretical and experimental foundations in this paper, we can extend QIREN to these applications in future work and QIREN is anticipated to yield superior results with fewer parameters in these domains. Simultaneously, we have found a suitable application scenario for quantum machine learning. This will stimulate further practical and innovative research within the quantum machine learning community.

\section*{Acknowledgements}
This work is supported in part by the Natural Science Foundation of China (grant No. 62276188), TJU-Wenge joint laboratory funding.
\section*{Impact Statement}
This paper presents work whose goal is to advance the field of Machine Learning. There are many potential societal consequences of our work, none which we feel must be specifically highlighted here.


\bibliography{example_paper}
\bibliographystyle{icml2024}

\newpage
\appendix
\onecolumn
\setcounter{table}{0} 
\setcounter{figure}{0}
\setcounter{equation}{0}
\renewcommand{\thetable}{A\arabic{table}}
\renewcommand{\thefigure}{A\arabic{figure}}
\renewcommand{\theequation}{A\arabic{equation}}

\section{Preliminaries of Quantum Circuit}\label{qc}
In quantum computing, information is often carried by qubits over Hilbert space. A pure quantum state consists of one or more qubits and is usually represented by Dirac’s notation, which denotes a unit vector $\textbf{v}$ as a ket $|v\rangle$ and its conjugate transpose $\textbf{v}^\dagger$ as a bra $\langle v|$. The inner product between $|v\rangle$ and $|u\rangle$ is denoted as $\langle u|v\rangle$, and the outer product is $|u\rangle\langle v|$. The evolution of a quantum state $|v\rangle$ is accomplished by sequentially applying quantum gates on it, i.e. $\left|v^{\prime}\right\rangle=U_{K}...U_{2}U_{1}|v\rangle$, where $U_{k}$ is the unitary matrix representing the quantum gate and $\left|v^{\prime}\right\rangle$ is the quantum state after evolution. Common single-qubit gates are as follows:
\begin{equation}
    \begin{split}
    H&: = \frac{1}{\sqrt{2}} \left[\begin{array}{cc}1 & 1 \\1 & -1\end{array}\right], I: =  \left[\begin{array}{cc}1 & 0 \\0 & 1\end{array}\right],\\
    X&: = \left[\begin{array}{cc}0 & 1 \\1 & 0\end{array}\right], 
    R_X(\theta) :=\left[\begin{array}{cc}\cos{\frac{\theta}{2}} & -i\sin{\frac{\theta}{2}} \\-i\sin{\frac{\theta}{2}} & \cos{\frac{\theta}{2}}\end{array}\right],\\
    Y&: = \left[\begin{array}{cc}0 & -i \\i & 0\end{array}\right],R_Y(\theta) :=\left[\begin{array}{cc}\cos{\frac{\theta}{2}} & -\sin{\frac{\theta}{2}} \\\sin{\frac{\theta}{2}} & \cos{\frac{\theta}{2}}\end{array}\right],\\
    Z&: = \left[\begin{array}{cc}1 & 0 \\0 & -1\end{array}\right],R_Z(\theta) :=\left[\begin{array}{cc}1 & 0 \\0 & e^{i\theta}\end{array}\right],
    \end{split}
\end{equation}
where $H$ denotes the Hadamard gate, $X, Y, Z$ denote the Pauli gates, $R_X(\theta),R_Y(\theta),R_Z(\theta)$ denote the rotation gates. A common multi-qubit gate is $CNOT$ gate:
\begin{equation}
CNOT: = \left[\begin{array}{cccc}1 & 0 & 0 & 0 \\0 & 0 & 0 & 1 \\ 0 & 0 & 1 & 0 \\0 & 1 & 0 & 0 \end{array}\right].
\end{equation}

In a quantum circuit, the initial quantum state is generally $|0\rangle^{\otimes N}$,
and after applying a sequence of quantum gates, the measurement will be used to convert quantum information into classical information. For instance, we can design quantum measurements to obtain the expectation $\langle v|O| v\rangle$ of the quantum state $|v\rangle$ about an observable $O$.

\section{Additional Explanation on Claim 2}\label{claim2}
The role of the Linear layer in QIREN is actually an improved version of the exponential encoding scheme \cite{shin2023exponential}. We will next explain the role of the Linear layer and conclude with a comparison with the exponential encoding scheme that illustrates the advantages of the Linear layer encoding scheme.

If we ignore the bias term, the Linear layer essentially maps the input $x$ to $(w_1x,\ldots,w_dx)$. Considering the form of the encoding gate as $S(x)= \otimes_{q=1}^{d} \mathrm{e}^{-\mathrm{i} (w_{q} x) Z / 2}=\otimes_{q=1}^{d} \mathrm{e}^{-\mathrm{i}  x(w_{q} Z / 2)}$, the specific form of $H$ is as follows:
\begin{equation}
H= \sum_{q=1}^{d} w_{q} Z^{(q)}/ 2.
\end{equation}
In the exponential encoding scheme, $H= \sum_{q=1}^{d} \beta_q Z^{(q)}/ 2.$ The difference between the two methods is that $w_q$ is a learnable parameter while $\beta_q$ is a hyperparameter that needs to be manually tuned. In the absence of degeneracy, $H$ can have $2^d$ distinct eigenvalues. Thus, the spectrum can contain at most $4^d - 2^d + 1$ distinct frequencies (including zero). But if in the absence of the Linear layers i.e. $w_q=1$, then there will be only $d+1$ distinct eigenvalues and the size of the spectrum $\Omega=\left\{\lambda_{k}-\lambda_{j} \right\}$ will be limited to $2d+1$. Next we derive a theoretical upper bound on the spectrum that can be extended by the Linear layer, where the derivation follows the idea in the exponential encoding scheme \cite{shin2023exponential}. 

With $H= w_{1} Z / 2$, we have $\Omega^{(1)} = \{-w_1,0,w_1\}$ as the frequency spectrum. If we append one more encoding gate to $H$,we have $\Omega^{(2)} = \{-w_2 - w_1,-w_2,-w_2 + w_1,-w_1,0,w_1,w_2 - w_1,w_2,w_2 + w_1\}$. More generally, let $\Omega^{(k)}$ as the $k$-th frequency spectrum as a result of using $k$ encoding gates. Then $\Omega^{(k)}$ has all elements from $\Omega^{(k-1)}$, along with the new ones generated by adding $\pm w_k$ to all elements of $\Omega^{(k-1)}$:
\begin{equation}
\Omega^{(k)}=\left\{\Omega^{(k-1)}-w_{k}, \Omega^{(k-1)}, \Omega^{(k-1)}+w_{k}\right\}.
\end{equation}
As $\Omega^{(k)}$ is always symmetric about zero, to generate a maximally non-degenerate frequency spectrum, the inequality
\begin{equation}\label{eqclaim2}
\max \left\{\alpha \in \Omega^{(k-1)}\right\}<w_{k}-\max \left\{\alpha \in \Omega^{(k-1)}\right\}
\end{equation}
is to be satisfied, where we can easily deduce that $\max \left\{\alpha \in \Omega^{(k-1)}\right\}=\sum_{j=1}^{k-1} w_{j}$. When Eq. (\ref{eqclaim2}) is satisfied, the spectrum can be extended from $2d+1$ to the theoretical upper bound $3^d$.

Finally, we compare the exponential encoding scheme \cite{shin2023exponential} with our Linear layer encoding scheme. First, exponential encoding scheme can always reach the theoretical upper bound of the extension of the spectrum by manually selecting the hyperparameters. However, when faced with signals with different spectrum, manual tuning obviously lacks flexibility. Second, after the neural network layers are stacked, the spectrum of the function to be fitted in each layer is unknown and the optimal hyperparameters cannot be manually selected. Our Linear layer encoding scheme, on the other hand, is learnable and utilizes gradient descent to achieve a suitable trade-off between reducing spectrum redundancy and covering the signal frequency. This allows us to implement an end-to-end training process without the need to manually tune the hyperparameters. Thus our approach is more practical and can be applied to complex real-world tasks.

\section{Additional Implementation Details}\label{imp}
\subsection{Signal Representation}
We use MSE as the loss function and use Adam optimizers with the parameters $\beta_1=0.9$, $\beta_2=0.999$ and $\epsilon=1e-8$. The models are trained for 600 epochs for Astronaut, 300 epochs for Camera, Coffee or sound. 

\textbf{ReLU-based MLP} includes one input linear layer, one output linear layer, and six hidden linear layers with a hidden dimension of 10. Except for the output layer, the output of each linear layer will pass through a BatchNorm layer and then the activation function.

\noindent \textbf{Tanh-based MLP} is set up the same way as the ReLU-based MLP.

\noindent \textbf{ReLU-based MLP with Random Fourier Features} is set up the same way as the ReLU-based MLP, except that the input layer is replaced with an RFF layer which consists of a non-trainable random parameter matrix followed by sine and cosine activation functions.

\noindent \textbf{SIREN} is set up the same way as the ReLU-based MLP, but instead of using BatchNorm layers, it employs a special initialization method and hyperparameter tuning to regulate the data distribution.

\noindent \textbf{QIREN} includes three Hybrid layers and one output linear layer. Each Hybrid Layer consists of a linear layer with a hidden dimension of 8, a BatchNorm layer and a quantum circuit with 8 qubits.

\subsection{Image Generation}
\begin{figure}[h]
\centering
\includegraphics[scale=0.42]{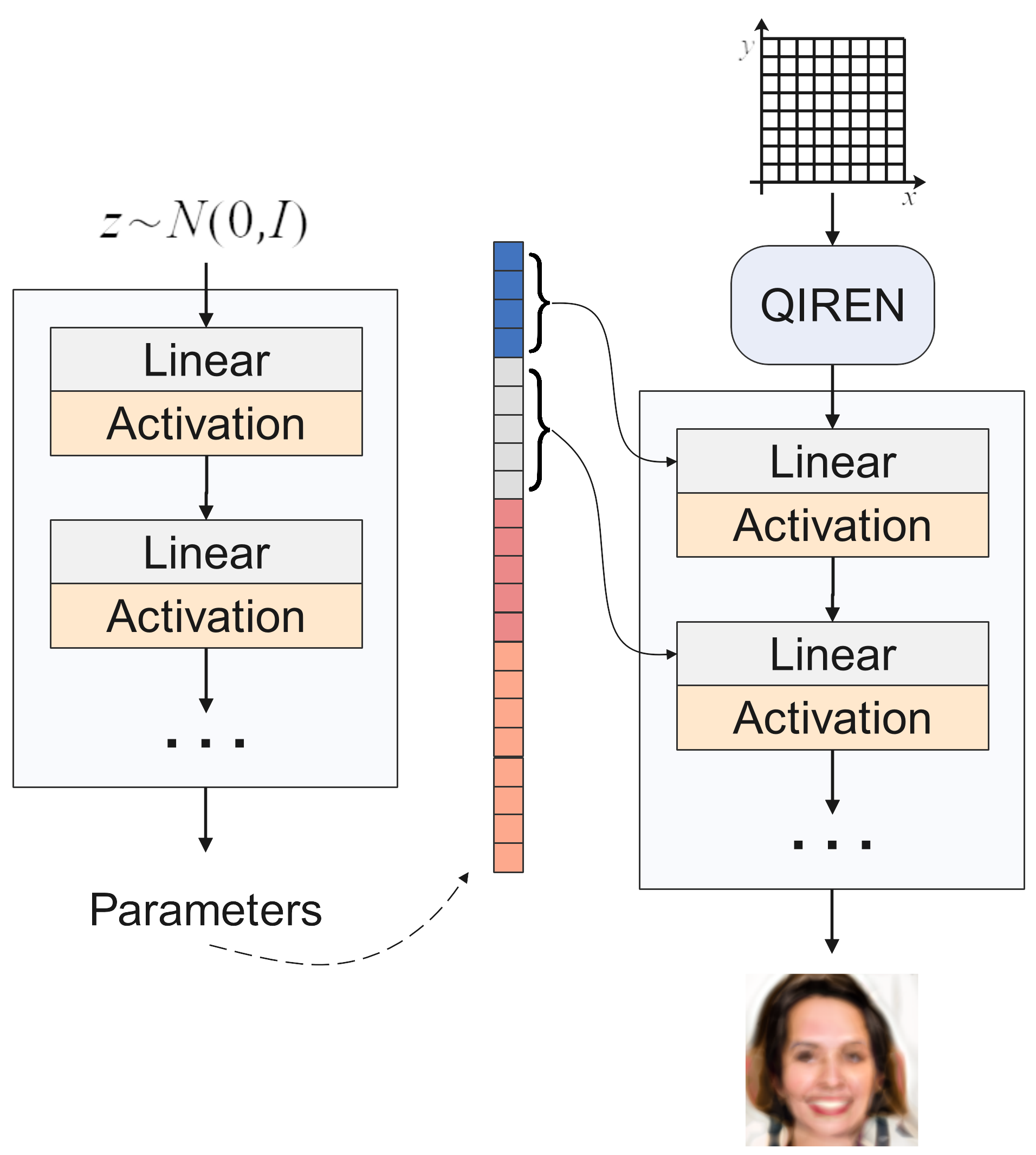}
\caption{QIREN-based generator.}
\label{fig8}
\end{figure}

\begin{figure}[!h]
\centering
\includegraphics[scale=0.8]{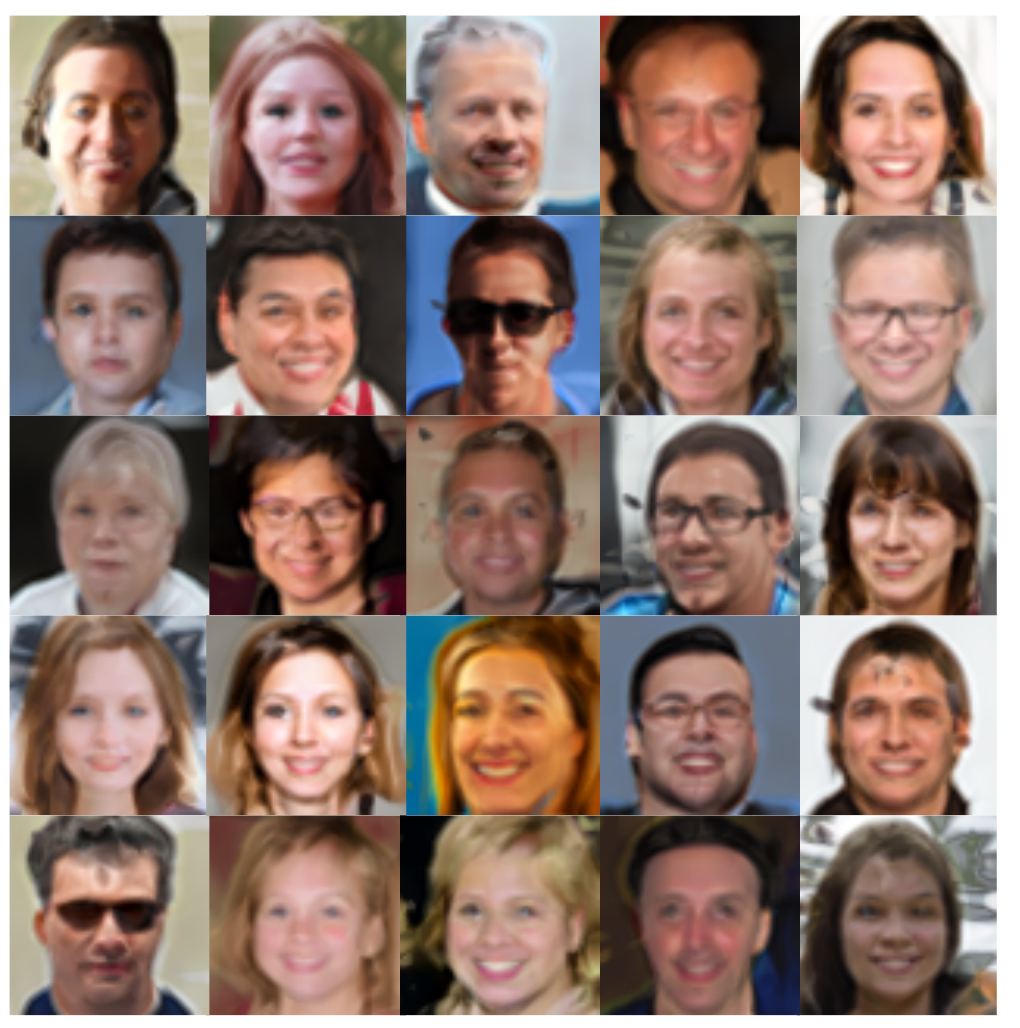}
\caption{Results of image generation.}
\label{fig18}
\end{figure}
We build upon the StyleGAN2 framework \cite{karras2020analyzing}, replacing a convolutional generator with our INR-based one as illustrated in Figure \ref{fig8}. For several classical INR-based generators, we directly generated all their parameters using a hypernetwork. However, for QIREN, based on our experiments, we found that the hypernetwork is unable to comprehend the quantum components, leading to poor performance when directly generating parameters. Therefore, we added a ReLU-based MLP after QIREN, and the hypernetwork only generates parameters for this ReLU-based MLP. In this setup, QIREN plays a role similar to that of RFF. All generators include a hypernetwork with eight hidden layers with a hidden dimension of 256. We use non-saturating logistic loss for training:
\begin{equation}
\begin{array}{cc}
     Loss_{G}=\frac{1}{m} \sum_{i=1}^{m} -\log \left(D\left(G\left(z^{(i)}\right)\right)\right),& \\
     Loss_{D}=\frac{1}{m} \sum_{i=1}^{m}\left[\log D\left(\mathbf{x}_i\right)+\log \left(1-D\left(G\left(\mathbf{z}_i\right)\right)\right)\right],&
\end{array}
\end{equation}
and use Adam optimizers with the parameters $\beta_1=0.0$, $\beta_2=0.98$ and $\epsilon=1e-8$. All models are trained for 6048 kimg. Additionally, we utilize a trick named factorized multiplicative modulation layer, which was proposed in previous works \cite{skorokhodov2021adversarial}, to alleviate the parameter burden of the hypernetworks. For more specific details about the architecture of the INR-based generators, please refer to the code.

\section{Results on Image Generation}\label{img_gen}
In Figure \ref{fig18}, we present the $32 \times 32$ pixel images generated by the QIREN-based generator. Next, we will demonstrate two exciting properties that set the QIREN-based generator apart from traditional convolutional generators: out-of-the-box superresolution and meaningful image-space interpolation.

\begin{figure}[h]
\centering
\includegraphics[scale=0.5]{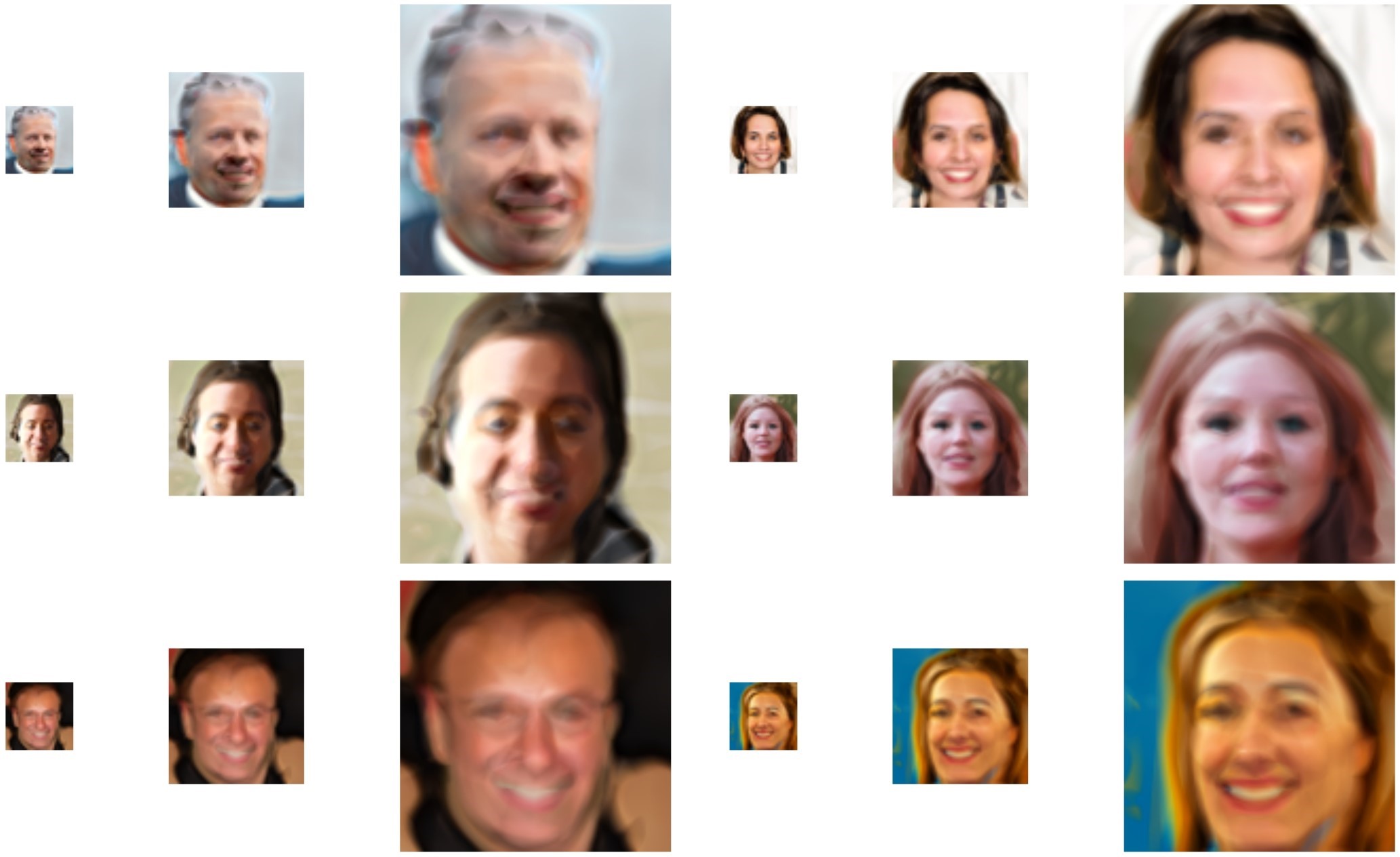}
\caption{Superresolution of generated images.}
\label{fig16}
\end{figure}

Our QIREN-based generator is able to produce images of higher resolution than it was trained on. For this, we evaluate our model on a denser coordinates grid. Specifically, we input coordinate grids of size $64 \times 64$ and $128 \times 128$ separately into the QIREN-based generator, and the results are shown in Figure \ref{fig16}.

\begin{figure}[h]
\centering
\includegraphics[scale=0.7]{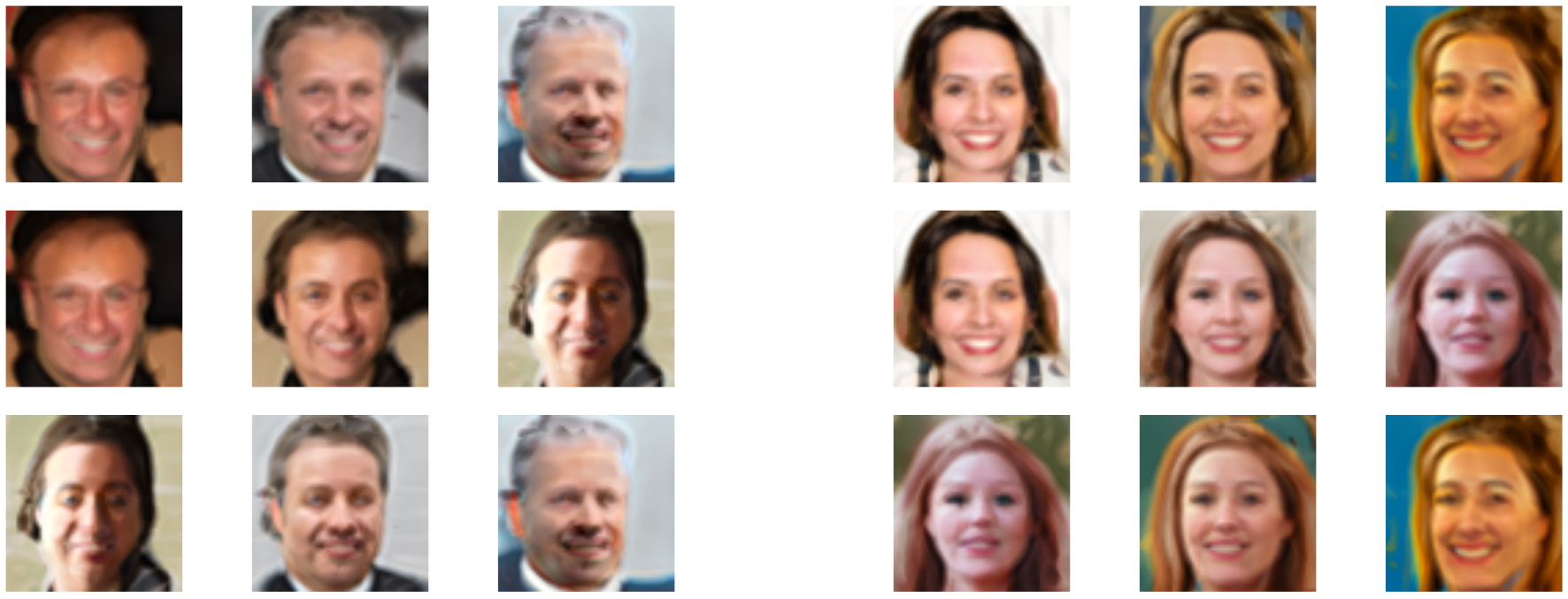}
\caption{The intermediate image is the result of interpolating between two images on the left and right. To interpolate between $G(\theta_1)$ and $G(\theta_2)$ , we compute interpolation parameters $\theta = \alpha\theta_1+ (1 - \alpha)\theta_2$ and evaluate $G(\theta)$ for the provided coordinates grid. The interpolated images can also achieve superresolution.}
\label{fig17}
\end{figure}

As is well known, image-space interpolation often exhibits poor performance. However, in Figure \ref{fig17}, we demonstrate that the images generated by the QIREN-based Generator can be interpolated reasonably.

\begin{figure}[hb]
\centering
\includegraphics[scale=0.32]{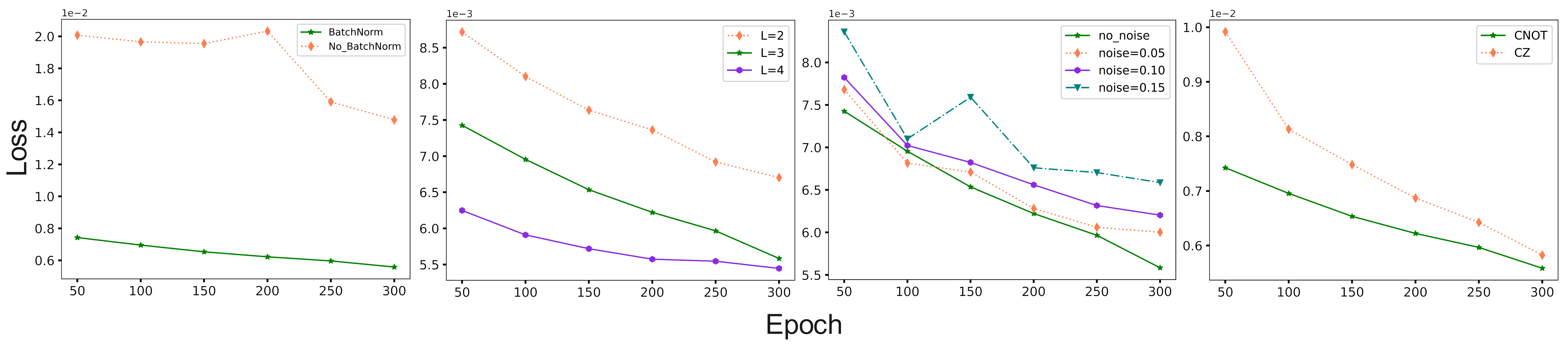}
\caption{Loss functions of ablation experiments on the sound representation task.}
\label{fig9}
\end{figure}

\begin{figure}[h]
\centering
\includegraphics[scale=0.65]{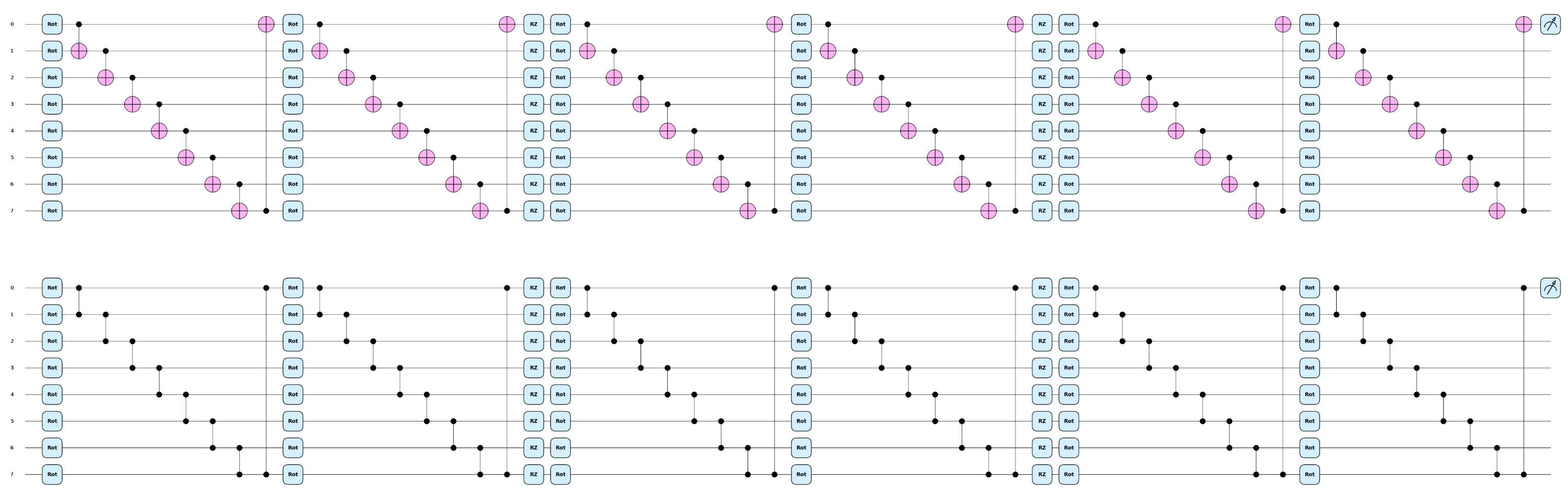}
\caption{Circuit diagrams for the two ansatz.}
\label{fig19}
\end{figure}

\section{Ablation Analysis}\label{aa}
\subsection{Signal Representation}
We conducted four ablation experiments on the sound representation task and presented the results in Figure \ref{fig9}. First, we ablate the importance of the BatchNorm layer. The loss function value of QIREN with BatchNorm consistently remained lower than the one without BatchNorm. This finding demonstrates that the role of the activation function has been taken over by the quantum circuit. As a result, the BatchNorm layer, which is typically placed before the activation function in classical models to deal with vanishing gradient problems, can also be applied to our quantum model to effectively promote fast convergence.

Next, we investigated the influence of the number of data re-uploading $L$, also known as circuit depth, on the representation capacity. In the original version of QIREN, $L$ was set to 3. It can be observed that within our experimental range, as $L$ increased, the representation capability of QIREN was strengthened.

Considering the real-world scenario of running on a quantum computer, we investigated the impact of noise on QIREN. We introduced noise by applying an $R_X(\theta)$ on each qubit before the measurement, where $\theta \sim \text{U}(0, \text{noise})$ and $\text{noise}=0.05, 0.10, 0.15$. It can be observed that noise does impact the performance of QIREN, but even with the presence of noise, QIREN demonstrates remarkable signal representation capability comparable to SIREN and ReLU-based MLP with RFF.

Finally, we investigated the impact of the quantum circuit ansatz on performance. In the vanilla QIREN, we used $CNOT$ gates to provide quantum entanglement. We attempted to replace them with $CZ$ gates, and both ansatz showed similar performance. The circuit diagrams for both ansatz are shown in Figure \ref{fig19}.

We performed an ablation analysis of the Linear layer on the image representation task, and as can be seen in Table \ref{tb4}, the model fits the image poorly after removing the Linear layer. This is because the Linear layer serves to extend and adjust the spectrum, allowing QIREN to capture most of the important frequency bands of the signal with a small amount of quantum resources, as analyzed in Claim 2 and Appendix \ref{claim2}. However, a limitation of this ablation analysis is that after removing the Linear layer, QIREN converts to a pure quantum circuit. The current limitations of hardware devices restrict us to training only a lightweight quantum circuit, with the number of parameters involved not being on the same order of magnitude as in QIREN. This is also the reason why we adopted a hybrid quantum neural network: to obtain quantum advantages in real-world tasks with limited quantum resources. Pure quantum circuits are often limited to simple synthetic tasks.
\begin{table}[h] 
\centering
\begin{tabular}{c|cccc} 
\toprule 
Method&Astronaut&Camera&Coffee&$\#$params\\ 
\midrule
Prue quantum circuit&71.1&51.9&73.2&72\\
QIREN&4.0&1.1&1.5&657\\
\bottomrule 
\end{tabular}
\caption{MSE ($\times 10^{-3}$) of different models on the image representation task.}
\label{tb4}
\end{table}

\subsection{Image Generation}
We conducted ablation experiments on the FFHQ dataset, and the results are presented in Table \ref{tb3}. First, we employed the factorized multiplicative modulation technique \cite{skorokhodov2021adversarial}, which significantly reduced the number of model parameters while ensuring high-quality generated images. Next, we expanded the encoding dimension from 8 to 32 by increasing the number of quantum circuits. This expansion resulted in a substantial improvement in generation quality. We also experimented with increasing the number of data re-uploading and found that it enhanced the performance of the model. Finally, we tested the performance of the mainstream generator model, StyleGAN2 \cite{karras2020analyzing}.
In order to make a fair comparison, we adjusted the hyperparameters of StyleGAN2 to ensure that the total number of parameters in both models was approximately the same. While StyleGAN2 outperforms our QIREN-based generator in terms of image generation quality, it is important to note that superior image generation performance is not the primary advantage of the INR-based generator model. As shown in several studies \cite{skorokhodov2021adversarial,dupont2021generative} and Appendix \ref{img_gen}, their strength lies in unique properties that mainstream generator models do not possess.

\begin{table}[h] 
\centering
\begin{tabular}{l|cc} 
\toprule 
Method&FID&$\#$params\\ 
\midrule
QIREN &11.77&1.45M\\ 
+ Factorized Multiplicative Modulation&11.53&1.13M\\ 
+ Encoding Dimension&7.67&1.19M\\ 
+ Number of Data Re-uploading&6.88&1.19M\\ 
\midrule
StyleGAN2&5.14&1.38M\\
\bottomrule 
\end{tabular}
\caption{Results of ablation experiments on the FFHQ dataset.}
\label{tb3}
\end{table}

\end{document}